\documentclass[sigconf]{acmart}
\usepackage{balance}
\usepackage{multirow}
\usepackage{enumerate}
\usepackage{pifont}
\usepackage{listings}
\usepackage{algorithm}
\usepackage{algorithmicx}
\usepackage{amsmath}
\usepackage{xcolor}
\usepackage{enumitem}
\usepackage{tcolorbox}
\usepackage{etoolbox}

\definecolor{darkred}{RGB}{139,0,0}

\makeatletter
\AfterEndEnvironment{algorithm}{\let\@algcomment\relax}
\AtEndEnvironment{algorithm}{\kern2pt\hrule\relax\vskip3pt\@algcomment}
\let\@algcomment\relax
\newcommand\algcomment[1]{\def\@algcomment{\footnotesize#1}}
\renewcommand\fs@ruled{\def\@fs@cfont{\bfseries}\let\@fs@capt\floatc@ruled
  \def\@fs@pre{\hrule height.8pt depth0pt \kern2pt}%
  \def\@fs@post{}%
  \def\@fs@mid{\kern2pt\hrule\kern2pt}%
  \let\@fs@iftopcapt\iftrue}
\makeatother

\makeatletter
\def\@fnsymbol#1{\ensuremath{\ifcase#1\or \dagger\or \ddagger\or
   \mathsection\or \mathparagraph\or \|\or **\or \dagger\dagger
   \or \ddagger\ddagger \else\@ctrerr\fi}}
\makeatother

\newcommand{\stitle}[1]{\vspace*{0.15em}\noindent{\bf #1\/}}

\AtBeginDocument{%
  }

\setcopyright{acmlicensed}
\copyrightyear{2025}
\acmYear{2025}
\setcopyright{acmlicensed}\acmConference[KDD '25]{Proceedings of the 31st ACM SIGKDD Conference on Knowledge Discovery and Data Mining V.2}{August 3--7, 2025}{Toronto, ON, Canada}
\acmBooktitle{Proceedings of the 31st ACM SIGKDD Conference on Knowledge Discovery and Data Mining V.2 (KDD '25), August 3--7, 2025, Toronto, ON, Canada}
\acmDOI{10.1145/3711896.3736861}
\acmISBN{979-8-4007-1454-2/2025/08}

\begin{document}
% \begin{sloppypar}

\title{Blurred Encoding for Trajectory Representation Learning}

\author{Silin Zhou}
\affiliation{%
  \institution{University of Electronic Science and Technology of China}
  \city{Chengdu}
  \country{China}
}
\email{zhousilinxy@gmail.com}

\author{Yao Chen}
\authornote{Shuo Shang and Yao Chen are corresponding authors.}
\affiliation{%
  \institution{National University of Singapore}
  \country{Singapore}
}
\email{yaochen@nus.edu.sg}

\author{Shuo Shang}
\authornotemark[1]
\affiliation{%
  \institution{University of Electronic Science and Technology of China}
  \city{Chengdu}
  \country{China}
}
\email{jedi.shang@gmail.com}

\author{Lisi Chen}
\affiliation{%
  \institution{University of Electronic Science and Technology of China}
  \city{Chengdu}
  \country{China}
}
\email{lchen012@e.ntu.edu.sg}

\author{Bingsheng He}
\affiliation{%
  \institution{National University of Singapore}
  \country{Singapore}
}
\email{hebs@comp.nus.edu.sg}

\author{Ryosuke Shibasaki}
\affiliation{%
 \institution{LocationMind Inc.}
 \city{Tokyo}
 \country{Japan}
}
\email{shiba@locationmind.com}


\begin{abstract}
Trajectory representation learning (TRL) maps trajectories to vector embeddings and facilitates tasks such as trajectory classification and similarity search. State-of-the-art (SOTA) TRL methods transform raw GPS trajectories to grid or road trajectories to capture high-level travel semantics, i.e., regions and roads. However, they lose fine-grained spatial-temporal details as multiple GPS points are grouped into a single grid cell or road segment. To tackle this problem, we propose the \textbf{BLU}rred \textbf{E}ncoding method, dubbed \textbf{BLUE}, which gradually reduces the precision of GPS coordinates to create \textit{hierarchical patches} with \textit{multiple levels}. The low-level patches are small and preserve fine-grained spatial-temporal details, while the high-level patches are large and capture overall travel patterns. To complement different patch levels with each other, our BLUE is an encoder-decoder model with a pyramid structure. At each patch level, a Transformer is used to learn the trajectory embedding at the current level, while \textit{pooling} prepares inputs for the \textit{higher level} in the encoder, and \textit{up-resolution} provides guidance for the \textit{lower level} in the decoder. BLUE is trained using the trajectory reconstruction task with the MSE loss. We compare BLUE with 8 SOTA TRL methods for 3 downstream tasks, the results show that BLUE consistently achieves higher accuracy than all baselines, outperforming the best-performing baselines by an average of 30.90\%. Our code is available at \url{https://github.com/slzhou-xy/BLUE}.
\end{abstract}





% \keywords{Trajectory representation learning, GPS point, Hierarchy, Patch}

\maketitle

\begin{figure}[!t]
    \centering
    \includegraphics[width=\linewidth]{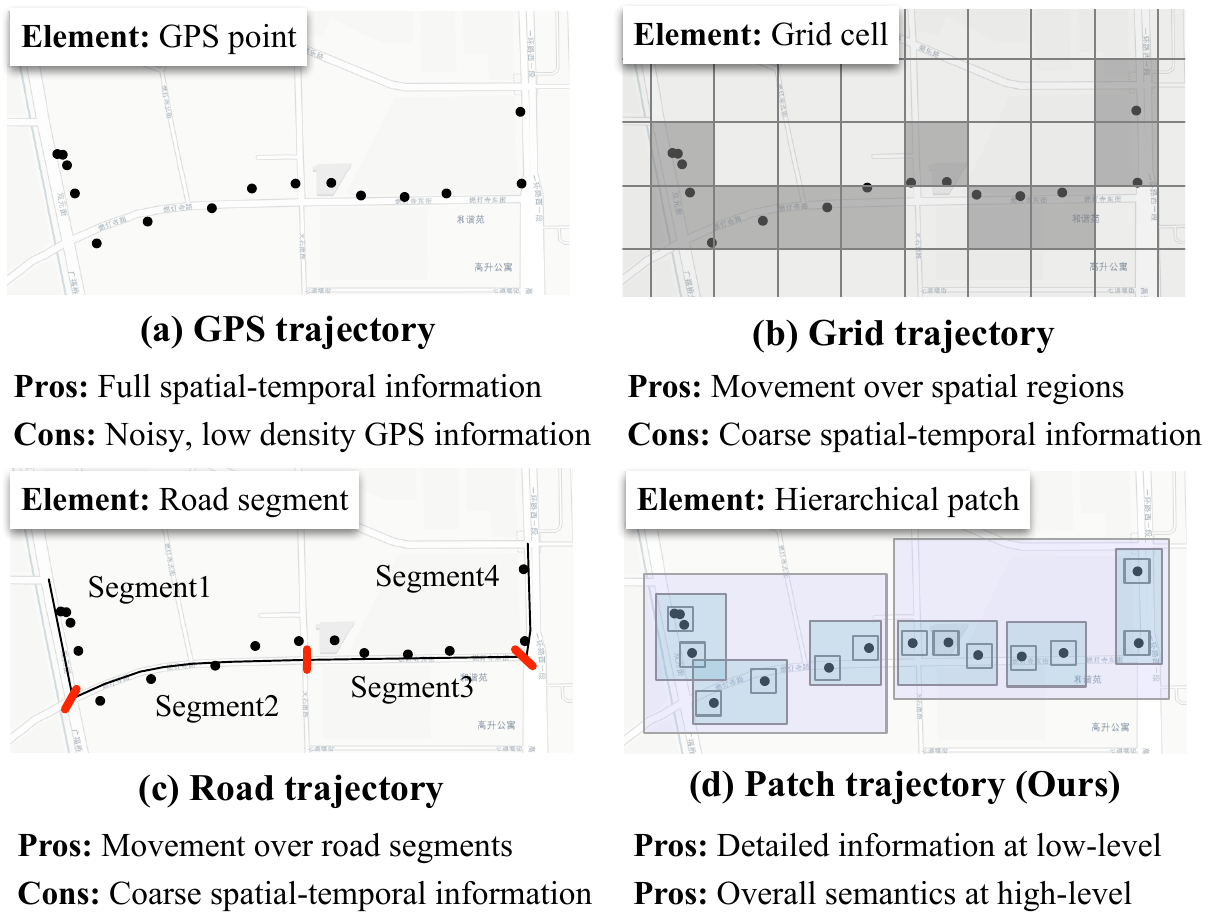}
    \caption{An illustration of different trajectory expressions.}
    \label{fig:intro}
\end{figure}

\section{Introduction}
With the popularization of GPS-enabled devices, human and vehicle movement trajectories can be easily collected. Analyzing these trajectories is crucial for many urban tasks such as traffic management~\cite{traffic_management}, business services~\cite{business_services}, and regional planning~\cite{region_emb}, which contribute to smarter cities~\cite{smart_city}. As a fundamental task for trajectory analysis, \textit{trajectory representation learning} (TRL) generates a vector for each trajectory while preserving the spatial-temporal travel semantics of the  trajectory~\cite{trl_survey,trajectory_survey}. By transforming variable-length trajectories into fixed-length vectors, TRL supports many downstream tasks, such as trajectory classification~\cite{TrajFormer,into_clustering}, travel time estimation~\cite{tte_intro,tte_intro2}, trajectory similarity computation~\cite{grlstm,intro_sim}, etc. For instance, trajectory similarity computation originally requires quadratic time w.r.t. trajectory length but becomes linear-time Euclidean distance computation with TRL~\cite{similarity_survey}.

\stitle{Existing works and their limitations.} The SOTA TRL methods use either grid trajectories~\cite{t2vec, trajgat,trajcl} or road trajectories~\cite{start,mmtec,jgrm} as inputs to tackle the irregularities and noises of GPS trajectories. Figure~\ref{fig:intro}(a) shows a raw GPS trajectory, which is a time-ordered sequence of GPS points. However, these points are spatially and temporally irregular due to variations in sampling intervals and positioning noise, and each GPS point carries limited spatial-temporal information and travel semantics~\cite{trajectory_survey}. Thus, directly modeling GPS trajectories yields low accuracy.
As shown in Figure~\ref{fig:intro}(b), a grid trajectory is a sequence of grid cells, where each cell corresponds to a region of the city and groups multiple GPS points. By grouping GPS points, grid trajectories reduce the influence of jitters and noises and can capture the high-level properties of regions (e.g., living or commercial areas)~\cite{trajectory_survey}. Figure\ref{fig:intro}(c) illustrates a road trajectory, which consists of road segments obtained by applying map-matching~\cite{map_matching_survey} on the raw GPS trajectories. Road trajectories represent object movement along the road network, with each segment corresponding to a street and grouping multiple GPS points.

Although grid and road trajectories benefit TRL by capturing high-level travel semantics, i.e., over regions or roads, they also suffer from three crucial limitations.
\ding{182} \textit{Limited resolution}. By grouping multiple GPS points into a grid cell or road segment, they capture high-level local information but lose fine-grained spatial-temporal details inherent to individual GPS points, which is essential for preserving basic spatial-temporal properties. In particular, a grid cell represents a geographical region and thus cannot assign precise timestamps to it. Similarly, a road segment usually only retains the average time of its GPS points~\cite{trajectory_survey}. The limited spatial-temporal resolution hinders tasks that require fine-grained information (e.g., travel time estimation). \ding{183} \textit{Poor generalization}. The grid cells and road networks differ for cities, and thus TRL models that use grid or road trajectories cannot generalize across different cities. However, a good generalization is favorable for reducing model training costs and handling cases when data is scarce. \ding{184} \textit{Poor robustness}. Grid-based TRL methods need to tune the size of grid cells for good accuracy~\cite{grid_size,similarity_survey}. The quality of the road trajectories depends on the map-matching algorithm and the precision of the road graph~\cite{fmm,graphmm}, and thus the accuracy of road-based TRL methods is also affected. Given the shortcomings of grid and road trajectory expressions, we pose the following research question:

\bgroup
\setlength{\leftmargini}{18pt} 
\setlength{\rightmargin}{18pt}
\begin{quote}
	\textit{Can we design a trajectory expression that attains both low-level spatial-temporal details and high-level travel semantics for accurate, general, and robust TRL?}
\end{quote}
\egroup

\noindent
\textbf{Our solution BLUE.} We propose the \textit{\textbf{blu}rred \textbf{e}ncoding} method, dubbed \textit{BLUE}, to transform each GPS trajectory into a \textit{patch trajectory}. As shown in Figure~\ref{fig:intro}(d), our patches have a \textit{hierarchy}, with each low-level patch covering a small area and a high-level patch encompassing multiple low-level patches. Moreover, each patch may contain a different number of GPS points. In particular, we drop the least significant decimals of the GPS coordinates and group the points that share a common prefix into one patch. By \textit{progressively} dropping more decimals for the patches in higher levels, we form the patch hierarchy. Using the low-level patches, our patch trajectory preserves the fine-grained details of the GPS trajectory, and using the high-level patches, the patch trajectory can capture high-level regions and travel semantics. Moreover, since high-level patches encompass low-level patches, there are opportunities for two adjacent patch levels to complement each other to improve the TRL model. Finally, the patch trajectory does not rely on grid size or map matching and thus is more general and robust.  

Based on the patch trajectory, we design the BLUE model, which has an encoder and decoder with a pyramid structure. In particular, the encoder progressively compresses and transforms low-level GPS trajectories into high-level patch trajectories with multiple model levels. Each model level involves two components, i.e., a \textit{Transformer} to capture global trajectory information at the current level, and a \textit{patch pooling} to capture local semantics of the patches and generate higher-level patch trajectory. Conversely, the decoder progressively reconstructs low-level GPS trajectories from high-level patch trajectories. Each level of the decoder encompasses a \textit{up-resolution} to recover the current-level trajectory from a higher level, and a \textit{Transformer} to refine the restored trajectory for improved accuracy. To handle variable patch lengths in patch pooling and different trajectory lengths during up-resolution, we design an attention-based method for patch pooling and use cross-attention for up-resolution. To train our BLUE, we employ the mean squared error (MSE) loss for reconstructing the raw GPS trajectories, which is more efficient than the cross-entropy loss (involving a large number of classes with the Softmax~\cite{word2vec}) used in grid-based and road-based TRL methods.

To evaluate BLUE, we compare it with 8 SOTA TRL methods for 3 popular tasks, i.e., travel time estimation~\cite{jgrm}, most similar trajectory search~\cite{trajcl}, and trajectory classification~\cite{start}. The results show that BLUE consistently outperforms all baselines across the tasks and datasets, with an average accuracy improvement of 71.57\%, 14.48\%, and 1.82\% for travel time estimation, most similar trajectory search, and trajectory classification, respectively. Ablation studies confirm the effectiveness of BLUE’s designs. Generalization tests show that BLUE performs much better than all baselines when it is trained on trajectories from one city but applied to another city, i.e., in the case of transferability.

To summarize, we make the following contributions:
\begin{itemize}[leftmargin=*]
    \item We observe that SOTA TRL methods use either grid or road trajectories, which have problems including limited resolution, poor generalization, and poor robustness.
    \item To tackle the problems of grid and road trajectories, we propose blurred encoding to generate patch trajectories, which have hierarchical levels by progressively dropping GPS decimals and thus capture both low-level details and high-level semantics.
    \item Based on patch trajectories, we design a TRL model named BLUE, which has an encoder and decoder with a pyramid structure, to augment adjacent levels of patch trajectories with each other.
\end{itemize}

\section{Related Work}
Early TRL studies~\cite{traj2vec,CSTRM} use the raw GPS trajectories. For example, Traj2vec~\cite{traj2vec} generates movement features from GPS points with an RNN-based Seq2Seq~\cite{seq2seq} model. CSTRM~\cite{CSTRM} introduces point-level and trajectory-level differences to learn the trajectory representation. However, these methods are usually designed for a specific task, i.e., Traj2vec for clustering and CSTRM for trajectory similarity computation, thus lacking generality. More importantly, the amount of information carried by a single GPS point is limited, and thus directly modeling it does not yield high accuracy~\cite{trajectory_survey}. To solve these problems, SOTA TRL methods focus on using the grid or the road trajectories derived from the GPS trajectories.

\begin{figure}[!t]
    \centering
    \includegraphics[width=\linewidth]{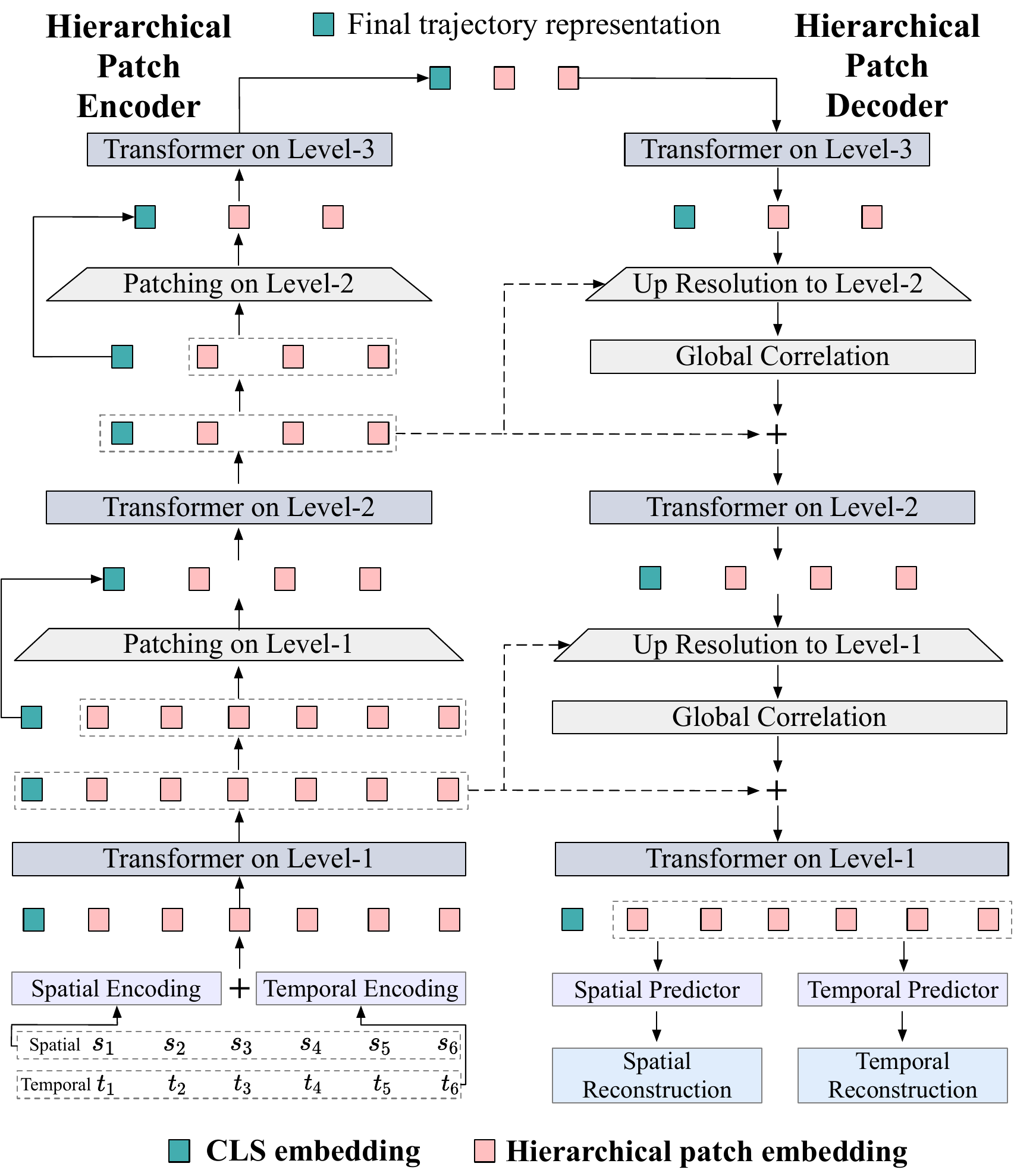}
    \caption{An overview of the BLUE model.}
    \label{fig:framework}
\end{figure}

\stitle{TRL with grid trajectories.}
To simplify and enhance GPS trajectories in free space, some methods~\cite{t2vec,e2dtc,trajgat,trajcl} use the grid cells to represent regions of a city map, and further group the multiple GPS points in the same region. Therefore, the GPS trajectory becomes the grid cell sequence, which introduces the region information. For example, t2vec~\cite{t2vec} uses an encoder-decoder model~\cite{seq2seq} to learn robust grid trajectory representations from low-quality data by a spatial proximity aware loss function and a cell pre-training algorithm. E$^2$DTC~\cite{e2dtc} proposes self-training to capture hidden spatial dependencies and learn grid trajectory representations. TrajCL~\cite{trajcl} uses contrastive learning~\cite{self_superviesd_survey} with various trajectory-specific augmentations to learn grid trajectory representations by introducing spatial and structural perspectives from GPS information. Although grouping points into a grid cell can reduce the GPS noise~\cite{trajectory_survey} and introduce region information, it changes the basic unit of the GPS trajectory and thus blurs the spatial-temporal information.

\stitle{TRL with road trajectories.}
Another way to reduce GPS noise and enhance the GPS trajectory is to use the map matching~\cite{fmm} algorithm to group multiple GPS points to a road segment to get the road trajectory. The road trajectory is continuous in the road network and can reflect the transition of object motion. Moreover, the road segment of the road network introduces the city topology and street information~\cite{trl_survey}. Therefore, most TRL studies~\cite{trembr,toast,pim,JCLRNT,lightpath,start} are based on the road trajectory. For example, Trembr~\cite{trembr} uses an RNN-based seq2seq model~\cite{seq2seq} to learn spatial-temporal information of the road trajectory. PIM~\cite{pim} introduces the graph embedding by node2vec~\cite{node2vec} to learn road segment embeddings and designs the mutual information maximization to learn the road trajectory representation. JCLRNT~\cite{JCLRNT} utilizes three types of contrastive loss, i.e., road-road, road-trajectory, and trajectory-trajectory, replacing detours and removing alternatives to learn the road trajectory representations. START~\cite{start} is a BERT-based~\cite{bert} model and incorporates travel semantics and temporal regularities with various data augmentations of the road trajectory. Although the road trajectory introduces additional information, it also changes the basic unit and has the problem of limited spatial-temporal information. 

Recently, some multi-modal methods~\cite{mmtec,jgrm} have been proposed to conduct TRL from multiple views. For example, JGRM~\cite{jgrm} learns trajectory representations by aligning the trajectory views of free space and road network jointly.
MMTEC~\cite{mmtec} proposes an attention-based discrete encoder and a NeuralCDE-based~\cite{cde} continuous encoder to extract the travel behavior and continuous spatial-temporal correlations of the trajectories.
However, they are still limited by using road trajectories. Different from existing TRL methods, our model provides a new trajectory expression, i.e., hierarchical patches, to capture the high-level semantics of the trajectories without losing the low-level details.
\section{Problem Definition} \label{sec:problem_formulation}
\noindent
{\bf Definition1 (GPS Trajectory).} A GPS trajectory $T$ is a time-ordered sequence of GPS points collected at a fixed or an unfixed sampling time interval by a GPS-enabled device. Each GPS point $g_i \in T$ is $(x_i, y_i, t_i)$, where $x_i$, $y_i$, and $t_i$ denote longitude, latitude, and timestamp, respectively.

\stitle{Definition2 (Trajectory Representation Learning).}
Given a set of GPS trajectories, trajectory representation learning (TRL) aims to learn a $d$-dimensional vector representation for each trajectory in the set. We expect learned representations to achieve high accuracy for downstream tasks, e.g., travel time estimation, trajectory classification, most similar trajectory search, etc. 
\section{The BLUE Model}
Figure~\ref{fig:framework} shows the overall structure of BLUE, which is an encoder-decoder model with a pyramid structure, consisting of four key modules: 1) Spatial-temporal Embedding: Transforms spatial-temporal GPS points into hidden embeddings. 2) Patch Encoder: Learns hierarchical spatial-temporal patch embeddings by progressively reducing GPS precision to create patches from high to low resolution. 3) Patch Decoder: Restores high-resolution trajectories from compressed low-resolution patches using cross-attention for length-insensitive restoration. 4) Spatial-temporal Reconstruction: Trains the model by reconstructing spatial-temporal GPS trajectories with MSE loss. For simplicity, multi-head attention and the [CLS] token are omitted in the Transformer formulas.

\begin{figure}[!t]
    \centering
    \includegraphics[width=\linewidth]{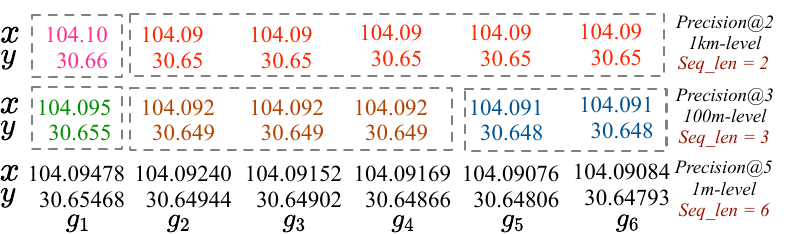}
    \caption{An illustration of blurred encoding by progressively rounding the decimals of the GPS points. The precision@5 is the raw GPS trajectory (i.e., level 1), while precision@3 and precision@2 are the patch trajectories at level 2 and level 3.}
    \label{fig:patch_method}
\end{figure}

\subsection{Blurred Encoding}
Before introducing the backbone network, we present blurred encoding. Inspired by patches~\cite{vit,patchtst}, we propose a trajectory-specific patch method based on different decimal precisions of GPS to progressively drop the least significant decimals of a GPS coordinate to decide the hierarchical patches. It generates patch trajectories to learn high-level travel semantics and preserves the GPS trajectories to learn detailed low-level spatial-temporal semantics.

Figure~\ref{fig:patch_method} illustrates the details of blurred encoding. Specifically, we consider three precision levels: precision@5 (level-1) with 1m resolution, representing the raw GPS trajectory; precision@3 (level-2) with 100m resolution, and precision@2 (level-3) with 1km resolution, denoting the patch trajectory. Precision@4 (10m resolution) is omitted as it shows no significant difference from precision@5. To transition from precision@5 to precision@3, we use a rounding method that groups GPS points with similar spatial semantics at precision@5 into the same patch at precision@3. This process transforms a GPS trajectory into a patch trajectory. For example, $g_2$, $g_3$, and $g_4$ at precision@5 are grouped into the same patch at precision@3. This grouping process is repeated to generate new patch trajectories at precision@2, capturing higher-level travel semantics. Notably, the input remains at precision@5, as lower precisions are not used directly but rather inferred through these mappings\footnote{To avoid noun confusion, we use the terms \textit{low-level} and \textit{high-level} to refer to high-resolution (i.e., high-precision) and low-resolution (i.e., low-precision), respectively, in the following sections.}.

This method leverages the semantics carried by GPS points to group low-level points into patches by blurring the GPS resolution. As a result, it offers several advantages:
\ding{182} \textit{Adaptive patch length}. The patch length adjusts dynamically based on GPS semantics and density by precision, making it flexible for trajectories of varying lengths. For example, a trajectory with many clustered GPS points will have a longer patch length, while a trajectory with fewer, widely spaced points may have a patch length of 1.
\ding{183} \textit{Semantic consistency}. GPS points within a patch share consistent low-level semantics, representing meaningful local properties. This reduces the spatial-temporal inconsistencies in the raw GPS trajectory and avoids truncating movement semantics.
\ding{184} \textit{Hierarchical representation}. Patch trajectories at different levels reveal unique information, reflecting hierarchical semantics. For instance, level-1, i.e., precision@5, retains basic spatial-temporal details, level-2, i.e., precision@3, captures local movement behaviors like turns, and level-3, i.e., precision@2, provides insights into long-range overall patterns. Moreover, the patch trajectory length at high-level is also greatly reduced, thus improving efficiency.

Compared to patches in computer vision (CV) and time series forecasting (TSF), our blurred encoding solves two issues in trajectories. \ding{182} \textit{Variable length}. In CV, batches of images with the same resolution of 512$\times$512 can be converted into patch sequences of length 64 using a fixed patch size of 64$\times$64~\cite{vit}. Similarly, in TSF, batches of sequences with a fixed length of 128 can be transformed into patch sequences of length 16 using a patch size of 8~\cite{patchtst}. However, trajectory batches vary significantly in length, ranging from just a few points to several hundred, making it impractical to define a fixed patch size for them. \ding{183} \textit{Local semantic}. In CV, a patch captures local spatial semantics, while in TSF, a fixed patch length (e.g., 12 for a 5-minute sampling rate) represents one hour of temporal semantics. However, for trajectories, a fixed patch length struggles to capture meaningful movement behavior, often segmenting dense clusters or continuous points, disrupting movement semantics.

\subsection{Spatial-temporal Embedding} 
The spatial-temporal embedding layer aims to convert coordinates and timestamps of the GPS trajectory, i.e., level-1, into spatial-temporal hidden embeddings. We design two encoding modules to achieve this: spatial encoding and temporal encoding.

\stitle{Spatial Encoding.} The raw GPS point $g_i$ only has limited spatial information, i.e., the coordinate. To enrich spatial information, we introduce additional spatial relations within a trajectory. In particular, for a GPS point $g_i$, we compute its \textit{forward distance} and \textit{forward azimuth} to the next GPS point $g_{i + 1}$, and its \textit{backward distance} and \textit{backward azimuth} to the previous GPS point $g_{i - 1}$ of a trajectory. If $g_i$ is the first point of a trajectory, the backward distance and backward azimuth are zero, and if $g_i$ is the last point of a trajectory, the forward distance and forward azimuth are zero. Therefore, combining the min-max normalized latitude and longitude information, we can obtain a spatial context vector $s_i \in \mathbb{R}^6$ for $g_i$ in a trajectory. Then we use a linear layer as the spatial encoding layer to transform $s_i$ to the spatial embedding as follows:
\begin{equation}\label{eq:spatial_emb}
\mathbf{e}_i^s = \mathrm{Linear} (s_i) \in \mathbb{R}^d,
\end{equation}

\stitle{Temporal Encoding.} The raw GPS time information of a GPS point $g_i$ is a timestamp, including the year, month, day, hour, minute, and second. However, the gap between these values is large. For example, the period of a year is usually 365, and the period of a second is 60, resulting in inconsistent input data ranges. To solve this issue, we use \textit{day-of-year}, \textit{day-of-month}, \textit{day-of-week}, \textit{hour-of-day}, \textit{minute-of-hour}, and \textit{second-of-minute}, to transform each value to the same range $[0, 1]$, we further subtract 0.5 to keep the range between $[-0.5, 0.5]$ of each value. Therefore, we obtain a temporal context vector $t_i \in \mathbb{R}^6$ of a GPS point $g_i$. Then we encode temporal information as follows:
\begin{equation}\label{eq:temporal_emb}
\mathbf{e}_i^{t} = \mathbf{W}_1 t_i \parallel \sin(\mathbf{W}_2 t_i), 
\end{equation}
where $\mathbf{W}_1, \mathbf{W}_2 \in \mathbb{R}^{\frac{d}{2} \times 6}$ are used to learn temporal context information, $\sin(\cdot)$ is further used to learn time periodicity information, $\parallel$ is the concatenation operation, and $\mathbf{e}_i^{t} \in \mathbb{R}^d$. Therefore, the final input embedding is $\mathbf{e}_i = \mathbf{e}_i^s + \mathbf{e}_i^t$ for a GPS point $g_i$. 

\begin{figure}
    \centering
    \includegraphics[width=\linewidth]{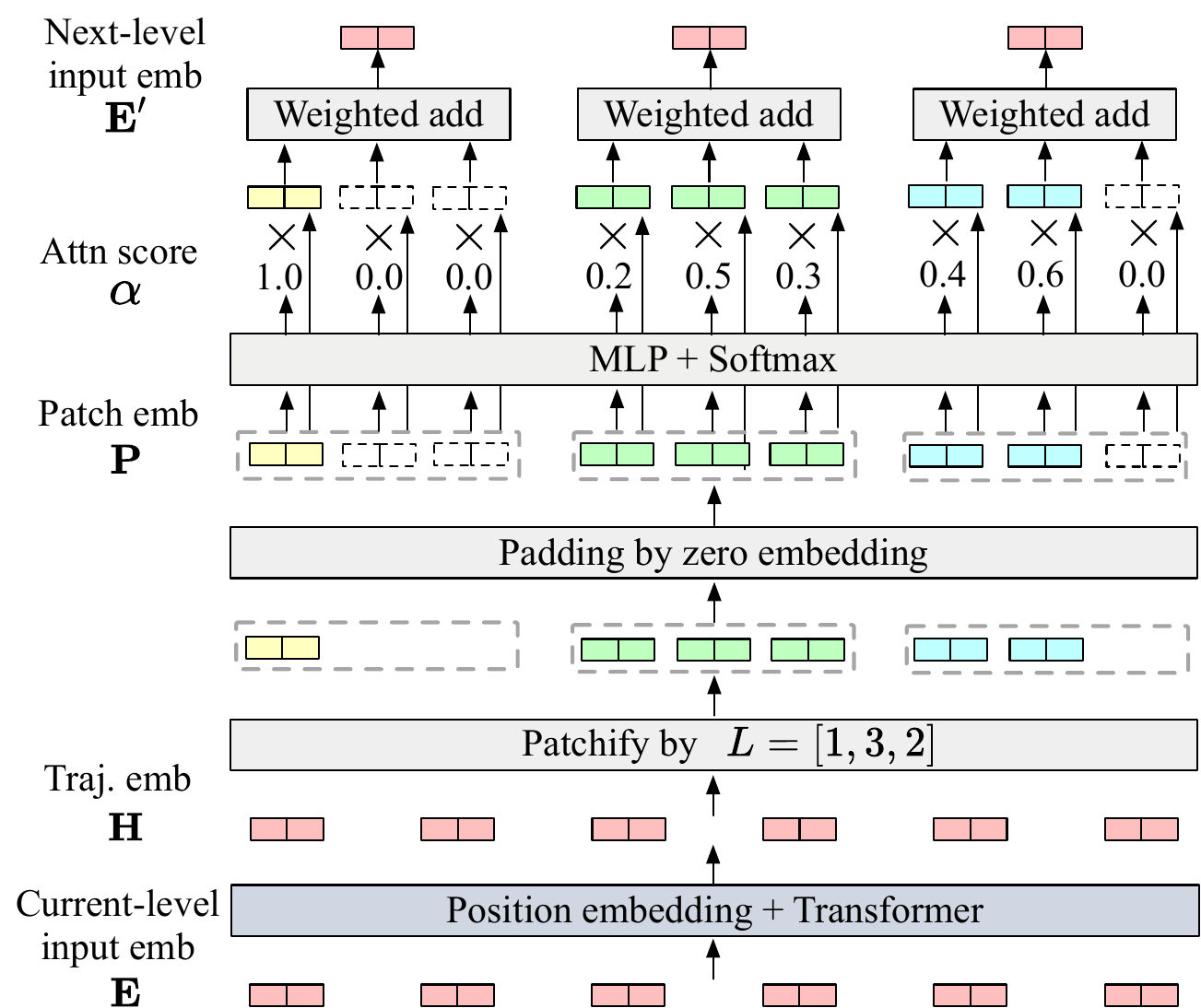}
    \caption{The patch encoder of BLUE from level-1 to level-2, the embedding dimension is 2, and the [CLS] token is omitted for simplicity. The encoder from level-2 to level-3 is similar.}
    \label{fig:pipeline}
\end{figure}

\subsection{Pyramid Encoder-decoder Structure}
Our backbone network is an encoder-decoder model with a pyramid structure. The encoder is used to learn the trajectory representation from low-level to high-level by compressing the trajectory with blurred encoding. The decoder aims to restore the low-level trajectory from the compressed high-level trajectory. 

\stitle{Patch Encoder.} The patch encoder alternates between the Transformer and blurred encoding. Figure~\ref{fig:pipeline} shows the pipeline of the encoder. The Transformer is used to capture global sequence information at the current-level. Blurred encoding by patching and pooling is to compress the trajectory embedding and generate the patch embedding with local semantics, then serve as the input for the next-level. In particular, we first add position encoding~\cite{transformer}, then use the Transformer on GPS trajectory embedding $\mathbf{e}_i$ to learn basic spatial-temporal information at level-1 as follows:
\begin{equation}\label{eq:transformer}
\begin{aligned}
    \mathbf{E} &= [ \mathbf{e}_1,  \mathbf{e}_2, ..., \mathbf{e}_{\vert T \vert} ],  \\
    \mathbf{H} &= \textrm{TransformerEncoder}(\mathbf{E}),
\end{aligned}
\end{equation}
where $\mathbf{H} \in \mathbb{R}^{\vert T \vert \times d}$ is trajectory embedding with basic spatial-temporal and global information at level-1. Then we blur it by patching $\mathbf{H}$ to the patch trajectory embedding as follows:
\begin{equation}\label{eq:patchify}
\begin{aligned}
    \mathbf{H} &= [\mathbf{h}_1,\mathbf{h}_2, ..., \mathbf{h}_{\vert T \vert}], \\
    L &= [l_1, l_2, ..., l_{\vert T_n \vert}], \\
     \mathbf{P} &= \textrm{Padding}(\textrm{Patchify}(\mathbf{H}, L)), \\
\end{aligned}
\end{equation}
where $l_i \in L$ is a scalar that denotes a patch length for $\mathbf{H}$ and $\sum_{i=1}^{\vert T_n \vert} l_i = \vert T \vert$, which can be obtained using blurred encoding when loading data. $\vert T_n \vert$ is the length of the patch trajectory of the next-level, i.e., level-2.
$\textrm{Patchify}(\cdot, \cdot)$ groups GPS point embeddings into a patch by using $L$. $\textrm{Padding}(\cdot)$ is used to keep the variable patch length the same. Therefore, $\mathbf{P} \in \mathbb{R}^{\vert T_n \vert \times M \times d}$, where $M$ denotes the maximum patch length.
To better explain Equation~\ref{eq:patchify}, we use a simple example, i.e., from level-1 to level-2, to illustrate it:

\underline{\textit{Example 1}}: The GPS trajectory $T = [g_1, g_2, ..., g_6]$ of level-1 is first learned to trajectory embedding $\mathbf{H} = [\mathbf{h}_1,\mathbf{h}_2, ..., \mathbf{h}_{6}]$ by using Equations~(\ref{eq:spatial_emb})-(\ref{eq:transformer}). $L = [1, 3, 2]$ records the patch length from level-1 to level-2, where $M=3$ and $\textrm{Sum}(L) = \vert T \vert$ (i.e., $\textrm{Sum}(L) = 6$). Then we use $\textrm{Patchify}(\mathbf{H}, L)$ to get patch embedding sequence $\mathbf{P} = [\mathbf{p}_1, \mathbf{p}_2, \mathbf{p}_3]$, where $\mathbf{p}_1 = [\mathbf{h}_1]$, $\mathbf{p}_2 = [\mathbf{h}_2, \mathbf{h}_3, \mathbf{h}_4]$, and $\mathbf{p}_3 = [\mathbf{h}_5, \mathbf{h}_6]$. We use padding embedding $\mathbf{h}^p$ to make each patch in $\mathbf{P}$ as the same length. Therefore, the final patch embeddings are $\mathbf{p}_1 = [\mathbf{h}_1, \mathbf{h}^p, \mathbf{h}^p]$, $\mathbf{p}_2 = [\mathbf{h}_2, \mathbf{h}_3, \mathbf{h}_4]$, and $\mathbf{p}_3 = [\mathbf{h}_5, \mathbf{h}_6, \mathbf{h}^p]$, i.e. $\mathbf{P} \in \mathbb{R}^{3 \times 3 \times d}$. 

To generate trajectory embeddings at the next-level by patch embedding $\mathbf{P}$, we propose an attention-based pooling method to eliminate the effects of padding embedding and provide more effective pooling results. Specifically, the attention score for an embedding $\mathbf{p}_{ij} \in \mathbf{p}_i$ in a patch $\mathbf{p}_i \in \mathbf{P}$ is computed as follows:
\begin{equation}
    \alpha_{ij} = \textrm{MLP}(\mathbf{p}_{ij}),
\end{equation}
where the MLP consists of a linear layer, a layer normalization~\cite{layernorm}, a relu activation~\cite{relu}, and another linear layer, executed sequentially. In Example 1, $\mathbf{p}_{11} \in \mathbf{p}_1$ is $\mathbf{h}_1$. If $\alpha_{ij}$ is computed by padding embedding $\mathbf{h}^p$, we set $\alpha_{ij}$ as \texttt{float(-inf)}. Then we softmax-normalize attention-score $\alpha_{ij}$ in a patch as follows:
\begin{equation}
    \alpha_{ij}^\prime = \frac{e^{\alpha_{ij}}}{\sum_{k=1}^{M}e^{\alpha_{ik}}},
\end{equation}
If $\alpha_{ij}=$ \texttt{float(-inf)}, then $\alpha_{ij}^\prime = 0$.
The next-level input embedding can be obtained by using a weighted addition as follows:
\begin{equation}\label{eq:fuse}
\begin{aligned}
    \mathbf{e}^{\prime}_i &= \sum_{j = 1}^{M}\alpha_{ij}^\prime \cdot \mathbf{p}_{ij}, \\
    \mathbf{E}^{\prime} &= [\mathbf{e}_1^{\prime},\mathbf{e}_2^{\prime}, ..., \mathbf{e}_{\vert T_n \vert}^{\prime}], \vert T_n \vert < \vert T \vert.
\end{aligned}
\end{equation}
$\mathbf{E}^{\prime}$ is the patch trajectory embedding of level-2. \textit{Noted that [CLS] token is ignored in patching of Equation(4)-(7), after patching, [CLS] token is prepended to patch trajectory embedding.}

We repeat Equations~(\ref{eq:transformer})-(\ref{eq:fuse}) to derive the patch trajectory embedding $\mathbf{H}^{\prime}$
in Transformer at level-2 with global correlation and new patch trajectory embedding $\mathbf{E}^{\prime \prime}$ at level-3. Then Equation~(\ref{eq:transformer}) is executed once again to learn the final compressed patch trajectory representation $\mathbf{H}^{\prime \prime}$ at level-3 to capture high-level travel semantics and global information. \textit{The [CLS] token of $\mathbf{H}^{\prime \prime}$ is considered as the final trajectory representation and used for downstream tasks}. Noted that the sequence length is reduced, so positional encoding is still applied before each use of the Transformer.

\stitle{Patch Decoder.} The patch decoder is architecturally symmetric to the patch encoder, alternating between the up-resolution operation and the Transformer.  The up-resolution operation is designed to restore the next low-level from the current high-level. The Transformer is employed to capture global correlations in the restored sequence, enhancing reconstruction capabilities. Notably, we first apply a Transformer to derive the new trajectory embedding $\widehat{\mathbf{H}}^{\prime \prime}$ from the final trajectory representation $\mathbf{H}^{\prime \prime}$, ensuring: 1) Symmetry in the architecture. 2) A deeper patch decoder structure.

Next, we employ an up-resolution operation to restore the next low-level from the current high-level, serving as the inverse process of patching in the patch encoder. In CV, up-resolution can be achieved using techniques like up-pooling~\cite{Uppooling}, transposed convolution~\cite{transposed_cnn}, etc. However, these approaches operate on fixed input and output sizes, such as transitioning from a low-resolution of 64$\times$64 to a high-resolution of 512$\times$512. In trajectory data, the actual lengths of trajectories are variable. For example, a batch of trajectories has a shape of $(b,l)$, where $b$ is the batch size (number of trajectories) and $l$ is the maximum trajectory length. In this batch, most trajectories are shorter than $l$, requiring padding to equalize their lengths. Due to this variability in length, up-resolution used in CV cannot be directly applied to trajectories. To clarify this, consider a simple example as follows:

\underline{\textit{Example 2}}: The shape of compressed trajectories at high-level is $(3,6)$, where 3 represents the batch size and 6 is the maximum length in this batch. However, the actual length of compressed trajectories is $[4,6,3]$. Similarly, the shape of expected restored trajectories at low-level is $(3,15)$, where 15 is the maximum length of restored trajectories in this batch. The actual length of expected restored trajectories is $[15,12,10]$. No direct length correspondence can be established between high-level and low-level trajectories.

To address the above issue, we introduce an up-resolution operation based on the cross-attention. Specifically, the compressed trajectory embeddings are used as input, while the trajectory embeddings from the patch encoder at the corresponding resolution of the expected restored trajectories are leveraged to assist in the restoration process. The operation of restoration from level-3 to level-2 is formulated as follows:
\begin{equation}\label{eq:dec_cross_attn}
    \widehat{\mathbf{E}}^{\prime} = \textrm{CrossAttn}(\textrm{query}=\mathbf{H}^{\prime}, \textrm{key}=\widehat{\mathbf{H}}^{\prime \prime}, \textrm{value}=\widehat{\mathbf{H}}^{\prime \prime}),
\end{equation}
where $\mathbf{H}^{\prime}$ is the query, which comes from the output of the Transformer at level-2 of the patch encoder. Key and value are $\widehat{\mathbf{H}}^{\prime \prime}$ comes from the output of the Transformer at level-3 of the patch decoder. Noted that $\mathbf{H}^{\prime}$ as the query because the output shape of the cross-attention matches the shape of the query and the length of $\mathbf{H}^{\prime}$ is exactly what we want to restore from $\widehat{\mathbf{H}}^{\prime \prime}$. Specifically, $\mathbf{H}^{\prime}$ only contributes to the computation of the attention matrix. The actual restored trajectory embedding is computed by weighted addition with $\widehat{\mathbf{H}}^{\prime \prime}$, ensuring that $\mathbf{H}^{\prime}$ is solely to assist reconstruction.

As the sequence length of restored $\widehat{\mathbf{E}}^{\prime}$ increases, we apply a self-attention to enable the restored trajectory representation to capture global information over the extended length as follows:
\begin{equation}\label{eq:dec_self_attn}
    \widebar{\mathbf{E}}^{\prime} = \textrm{SelfAttn}(\textrm{query}=\widehat{\mathbf{E}}^{\prime}, \textrm{key}=\widehat{\mathbf{E}}^{\prime}, \textrm{value}=\widehat{\mathbf{E}}^{\prime}),
\end{equation}
We once again utilize $\mathbf{H}^{\prime}$ from the patch encoder as the shortcut connection~\cite{resnet}, adding it with the restored trajectory embedding $\widebar{\mathbf{E}}^{\prime}$ as input to the Transformer, making more precise trajectory reconstruction at level-2:
\begin{equation}\label{eq:dec_transformer}
    \widehat{\mathbf{H}}^{\prime} = \textrm{TransformerEncoder}(\widebar{\mathbf{E}}^{\prime} + \mathbf{H}^{\prime}),
\end{equation} 
We repeat Equations(\ref{eq:dec_cross_attn})-(\ref{eq:dec_transformer}) to obtain trajectory embedding $\widehat{\mathbf{H}} \in \mathbb{R}^{\vert T \vert \times d}$ at level-1 using $\widehat{\mathbf{H}}^{\prime}$ of the patch decoder, and $\mathbf{H}$ of the patch encoder. Here, position encoding is discarded in the decoder because $\mathbf{H}^{\prime}$ and $\mathbf{H}$ carry position information in the patch encoder.

\subsection{Reconstruction Loss for Training}
We train our model by using spatial reconstruction loss and temporal reconstruction loss with the MSE loss function. Specifically, for $\widehat{\mathbf{h}}_i \in \widehat{\mathbf{H}}$, we first use two MLP networks to compute spatial and temporal predictions as follows:
\begin{equation}
        y_i^s = \textrm{MLP}_1(\widehat{\mathbf{h}}_i) \in \mathbb{R}^{6}, \ \ y_i^t = \textrm{MLP}_2(\widehat{\mathbf{h}}_i) \in \mathbb{R}^{6},
\end{equation}
then we compute spatial-temporal loss for trajectory $T$ as follows:
\begin{equation}
    \mathcal{L}^T = \sum_{i=1}^{\vert T \vert} (y_i^s - s_i)^2 + \sum_{i=1}^{\vert T \vert} (y_i^t - t_i)^2.
\end{equation}
We average the loss over the $\mathcal{N}$ trajectories in a mini-batch to obtain the final reconstruction loss $\mathcal{L}$.

Unlike previous studies that only focus on spatial loss~\cite{start,mmtec,jgrm}, we incorporate a temporal loss to enhance the reconstruction accuracy of low-level trajectories. Moreover, MSE loss is more efficient compared to the masked language model (MLM) loss~\cite{bert} used in the grid- and road-based methods~\cite{start,jgrm}, as MLM relies on classification with cross-entropy. Given that road segments and grids typically number in the tens of thousands, the Softmax operation in cross-entropy becomes highly inefficient~\cite{word2vec}.

Due to page limit, we analyze the complexity of BLUE in Appendix~\ref{sec:complexity}, and the results show that the overall complexity of BLUE is $O(L_1n_1^2)$, where $L_1$ and $n_1$ are the Transformer layers and sequence length of level-1, respectively. In Section~\ref{exp:micro}, we show by experiment that BLUE is competitive in efficiency when compared with SOTA TRL methods.

\begin{table}[tbp]
\LARGE
	\centering
	\caption{Statistics of the experiment datasets.}
	\label{tab:dataset}
	\resizebox{\linewidth}{!}{
		\begin{tabular}{l|c|c|c|c|c}\toprule
		\textbf{Dataset} & \textbf{\#Traj.} &  \textbf{Avg. length} & \textbf{Area size} & \textbf{Start time} & \textbf{End time} \\
			\midrule
                Porto   & 1,133,593 & 4,158meters & 78.3km$^2$ & 2013/07/01 & 2014/06/30 \\
                Chengdu & 1,114,194 & 4,702meters & 708.6km$^2$ & 2014/08/03 & 2014/08/05 \\
			\bottomrule
		\end{tabular}
	}
\end{table}

\section{Experimental Evaluation} \label{section:exp}
We experiment extensively to answer the following questions:
\begin{itemize}[leftmargin=*]
    \item \textbf{RQ1}: How does BLUE's accuracy compare with state-of-the-art TRL methods for various downstream tasks? 
    \item \textbf{RQ2}: How do BLUE's designs, i.e.,  hierarchical patches, and different pooling methods contribute to accuracy? 
    \item \textbf{RQ3}: How accurate is BLUE for cross-dataset model transfer?
    \item \textbf{RQ4}: How efficient is BLUE for training and inference?
    \item \textbf{RQ5}: How do BLUE's hyperparameters affect accuracy?
\end{itemize}

\begin{table*}[!t]
% \LARGE
	\centering
	\caption{Accuracy of BLUE and the baselines for the 3 downstream tasks on the Porto dataset. The best-performing baseline is marked with an underline, and the bottom row is the relative improvement of BLUE over the best baseline.}
	\resizebox{0.95\linewidth}{!}{
		\begin{tabular}{l|ccc|ccc|cc}
			\toprule
			& \multicolumn{3}{c|}{Travel Time Estimation (Seconds)} 
			& \multicolumn{3}{c|}{Most Similar Trajectory Search}
                & \multicolumn{2}{c}{Trajectory Classification}
			\\
			& MAE$\downarrow$  & MAPE(\%)$\downarrow$   & RMSE$\downarrow$  
			& \ \ MR$\downarrow$   & \ \ \ HR@1$\uparrow$     & HR@5$\uparrow$ 
            & \ \ \ Micro-F1$\uparrow$  & Macro-F1$\uparrow$
			\\
			\midrule
			Traj2vec 
			& 159.533  & 28.349  &  172.453   
			& \ \ 24.677 & 0.366 & 0.556
                & 0.686  & 0.650
			\\ 
			TrajCL 
			& 64.731  & 11.776  & 89.798     
			& \ \ 3.543 & 0.785 & 0.885  
                & \underline{0.823}  & \underline{0.795}
			\\
			Trembr 
			& 92.665  & 18.021  & 121.967     
			& \ \ 8.345 & 0.582 & 0.713
                & 0.810  & 0.753
			\\
			PIM 
			& 105.987  & 22.091  & 132.983  
			& \ \ 16.453 & 0.462 & 0.587
                & 0.770  & 0.714
			\\
			JCLRNT 
			& 98.356  & 20.014  & 129.786     
			& \ \ 11.754 & 0.540 & 0.648
                & 0.802  & 0.742
			\\
			START 
			& 83.355  & 15.158  & 116.624       
			& \ \ 2.932 & 0.801 & 0.854  
                & 0.819  &  0.772
			\\
			MMTEC
			& 62.673  & 11.593  & 87.072     
                & \ \ 2.023 & 0.831 & 0.927
                & 0.819 & 0.784
			\\
                JGRM
			& \underline{57.776}  & \underline{10.865}  & \underline{82.533}     
                & \ \ \underline{1.832} & \underline{0.858} & \underline{0.949}
                & 0.813  & 0.775  
			\\
                \midrule
                \textbf{BLUE} 
			& \textbf{6.756}  & \textbf{1.142}  & \textbf{12.060} 
			& \ \ \textbf{1.136}  & \textbf{0.911} & \textbf{0.998}
                & \textbf{0.833}  & \textbf{0.796}
			\\
			\midrule
			\textbf{Imp.}
			& 88.31\%  & 89.49\%  & 85.39\%    
			& \ \ 37.99\% & 6.18\% & 5.16\%
                & 1.22\%  & 0.13\% 
			\\
			\bottomrule
		\end{tabular}
	}
	\label{tab:performance_on_porto}
\end{table*}

\begin{table*}[!t]
% \LARGE
	\centering
	\caption{Accuracy of BLUE and the baselines for the 3 downstream tasks on the Chengdu dataset. The best-performing baseline is marked with an underline, and the bottom row is the relative improvement of BLUE over the best baseline.}
	\resizebox{0.95\linewidth}{!}{
		\begin{tabular}{l|ccc|ccc|ccc}
			\toprule
			& \multicolumn{3}{c|}{Travel Time Estimation (Seconds)} 
                & \multicolumn{3}{c|}{Most Similar Trajectory Search}
			& \multicolumn{3}{c}{Trajectory Classification} 
			\\
			& MAE$\downarrow$  & MAPE(\%)$\downarrow$       & RMSE$\downarrow$  
			& \ \ MR$\downarrow$   & \ \ \  HR@1$\uparrow$         & HR@5$\uparrow$ 
                & \ \ F1$\uparrow$     & Accuracy$\uparrow$    & Precision$\uparrow$
			\\
			\midrule
			Traj2vec 
			& 162.653 & 32.734  & 187.973   
			& \ \ 33.345  & 0.347  & 0.503
                & \ \ 0.757  & 0.671  & 0.713
			\\
			TrajCL 
			& 79.383  & 15.614  & 117.153  
			& \ \ 6.287  & 0.664  &  0.766
                & \ \ 0.872  & 0.802  & 0.828
			\\
			Trembr 
			& 94.435  & 18.794  &  127.546  
			& \ \ 9.345  & 0.532  & 0.673
                & \ \ 0.842  & 0.788  & 0.812
			\\
                PIM 
			& 116.954  & 22.895  & 139.866   
			& \ \ 19.423  & 0.403  & 0.536
                & \ \ 0.805  & 0.743  & 0.763
			\\
			JCLRNT  
			& 102.323  & 21.088  & 134.768   
			& \ \ 13.532  & 0.492  & 0.636
                & \ \ 0.824  & 0.763  & 0.791
			\\
			START 
			& 85.741  & 16.163  & 119.917   
			& \ \ 3.447  & 0.710  & 0.828
                & \ \ \underline{0.884}  & \underline{0.826}  & \underline{0.857}
			\\
                MMTEC 
			& 69.983  & 15.012  & 95.642   
                & \ \ 2.406  & 0.762  & 0.907
                & \ \ 0.849  & 0.786  & 0.814
			\\
			JGRM 
			& \underline{66.453}  &\underline{13.456}  &  \underline{93.846}  
                & \ \ \underline{2.006}  & \underline{0.784}  & \underline{0.951}
                & \ \ 0.868  & 0.792  & 0.818
			\\
			\midrule
			\textbf{BLUE} 
			& \textbf{29.302}  & \textbf{5.428}  & \textbf{46.299}   
			& \ \ \textbf{1.441}  & \textbf{0.831}  & \textbf{0.983}
                & \ \ \textbf{0.904}  & \textbf{0.855}  & \textbf{0.874}
                \\
                \midrule
			\textbf{Imp.}
			& 55.91\%  & 59.66\%  & 50.66\%   
			& \ \ 28.17\%  & 5.99\%  & 3.36\%
                & \ \ 2.26\%  & 3.51\%  & 1.98\%
			\\
			\bottomrule
		\end{tabular}
	}
	\label{tab:performance_on_chengdu}
\end{table*}

\subsection{Experiment Settings}
\noindent{\bf Datasets.} We experiment on two real-world trajectory datasets, i.e., Porto\footnote{https://www.kaggle.com/c/pkdd-15-predict-taxi-service-trajectory-i} and Chengdu\footnote{https://www.pkbigdata.com/common/zhzgbCmptDetails.html}, which are widely used by previous TRL studies~\cite{start,mmtec,jgrm}. The proportions of training, validation, and testing data are set to [0.6, 0.2, 0.2] for both datasets. The statistics of the datasets are shown in Table~\ref{tab:dataset}. 
GPS points include latitudes, longitudes, and timestamps, which are sampled every 15 seconds and 30 seconds in Porto and Chengdu on average, respectively. Porto records 3 different travel modes for vehicles: 1) The trip is dispatched from the central; 2) The trip is demanded directly to a taxi driver on a specific stand; 3) Otherwise (e.g., a trip demanded on a random street). Chengdu includes whether a taxi was carrying a passenger. After preprocessing (in Appendix~\ref{sec:appendix_preprocessing}), we get 1,133,593 and 1,114,194 trajectories in Porto and Chengdu, respectively.

\stitle{Baselines.}
We compare BLUE with 8 state-of-the-art TRL methods, \textit{Traj2vec}~\cite{traj2vec}, \textit{TrajCL}~\cite{trajcl}, \textit{Trembr}~\cite{trembr}, \textit{PIM}~\cite{pim}, \textit{JCLRNT}~\cite{JCLRNT}, \textit{START}~\cite{start}, \textit{MMTEC}~\cite{mmtec}, \textit{JGRM}~\cite{jgrm}. Among the baselines, TrajCL is the best-performing grid-based method, START is the best-performing road-based method, and JGRM is the best-performing multi-modal method. More details are provided in Appendix~\ref{sec:appendix_baselines}.

\stitle{Implementation.} We implement BLUE on an Nvidia A6000 48GB GPU running Ubuntu 22.04LTS with PyTorch 2.4. The hidden embedding and trajectory representation dimensions are set to $d = 128$. The Transformer configuration includes: 2 layers at level-1, 4 layers at level-2, and 2 layers at level-3 for both the patch encoder and decoder. Each attention mechanism uses 4 heads with a dropout ratio of 0.1. We pre-train BLUE with the Adam~\cite{adam} optimizer, using a batch size of 256 and a learning rate of 1e-4. Following START~\cite{start}, all methods are pre-trained for 30 epochs. The maximum patch length $M$ is dynamically computed in a batch.

\stitle{Downstream Tasks and Accuracy Measures.} We chose three downstream tasks that are widely used in TRL: travel time estimation~\cite{JCLRNT}, trajectory classification~\cite{start}, and most trajectory similarity search~\cite{trajcl}. We do not include trajectory recovery~\cite{RnTrajRec} or trajectory prediction~\cite{MobTCast} as downstream tasks because grid-based and road-based methods treat these as classification problems, while GPS-based methods (including ours) treat them as regression problems, thus their evaluation metrics differ, making direct comparisons infeasible. More detailed settings are in Appendix~\ref{sec:appendix_downstream settings}.

We follow existing TRL works~\cite{start,mmtec,jgrm} to choose the same accuracy measures for these tasks. In particular, for travel time estimation, we report the mean absolute error (MAE), mean absolute percentage error (MAPE), and root mean square error (RMSE). For multi-class trajectory classification on the Porto dataset, we report Micro-F1 and Macro-F1. For binary trajectory classification on the Chengdu dataset, we report the F1-score, accuracy, and precision. For most similar trajectory search, we report the hit ratio at the top-k results, i.e., the HR@1, HR@5, and the mean rank (MR).

\begin{table}[!t]
   \LARGE
   \centering
   \caption{Accuracy of BLUE when disabling the key designs. We denote travel time estimation as TTE, trajectory classification as TC, and most similar trajectory search as MSTS.}
   \resizebox{\linewidth}{!}{
   \begin{tabular}{l|ccc|ccc}\toprule
     & \multicolumn{3}{c}{Porto} & \multicolumn{3}{c}{Chengdu} 
      \\\cmidrule(lr){2-4}\cmidrule(lr){5-7}  
      & MAE$\downarrow$  & MR$\downarrow$ & Micro-F1$\uparrow$ 
        & MAE$\downarrow$ & MR$\downarrow$  & F1$\uparrow$ 
           \\
        & (TTE)  & (MSTS) & (TC)  
        & (TTE)  & (MSTS) & (TC)    \\
        \midrule
	w/o P@2
        & 7.089 & 1.219 & 0.820 
	& 29.581 & 2.187 & 0.891   
        \\
	w/o P@3 
 	& 6.932 & 1.389 & 0.809 
	& 30.564 & 4.619 & 0.901   
        \\
        w/o P@5 
 	& 7.418 & 20.677 & 0.815 
	& 34.025 & 14.047 & 0.884  
        \\
        w/o Patch
	& 10.082  & 1578.228  & 0.763  
 	& 34.503  & 4322.761  & 0.878  
        \\
	w/ Min 
 	& 7.708 & 1.166 & 0.822 
	& 35.729 & 2.568 & 0.900  
        \\
        w/ Max 
 	& 7.054 & 1.161 & 0.822  
	& 30.735 & 1.878 & 0.902   
        \\
        w/ Mean 
        & 7.485 & 1.165 & 0.821 
	& 29.407 & 2.131 & 0.901   
        \\
	\midrule
	\textbf{BLUE} 
 	& \textbf{6.756} & \textbf{1.136} & \textbf{0.833} 
	& \textbf{29.302} & \textbf{1.441} & \textbf{0.904} 
        \\
	\bottomrule
   \end{tabular}
   }
   \label{tab:ablation}
   \vspace{-3mm}
\end{table}

\subsection{Main Results (RQ1)}
Table~\ref{tab:performance_on_porto} and Table~\ref{tab:performance_on_chengdu} compare the accuracy of BLUE with the 8 baselines. We make several observations as follows.

In detail, the improvements of BLUE over the best-performing baseline are often substantial. For the travel time estimation task, the performance gain ranges from a minimum of 50.66\% to a maximum of 88.31\%. Notably, on the Porto dataset, our travel time estimation achieves an MAE in the single digits (measured in seconds) and a remarkably low MAPE of just 1\%. The performance of the TTE task is highly correlated with the sampling rate. Unlike road-based and grid-based methods, which group multiple GPS points into a road segment or a grid cell, leading to the loss of significant spatial-temporal details, e.g., sampling information. BLUE’s bottom layer preserves the full GPS spatial-temporal information, resulting in superior performance. Compared to the Porto dataset, the performance on the Chengdu dataset is slightly lower due to its highly uneven sampling rate, which ranges from 3 seconds to 60 seconds. In contrast, the sampling rate of the Porto dataset is typically around 15 seconds, providing more consistency.

For the most similar trajectory search task, BLUE achieves improvements ranging from 3.36\% to over 37.99\%. Although the irregular spatial-temporal nature of GPS trajectories typically hinders similarity search performance, BLUE mitigates this issue through blurred encoding and pyramid structure, transforming GPS trajectories into patch trajectories. This process reduces spatial-temporal irregularities, while the high-level trajectory representations effectively compress trajectory information and leverage low-level fine-grained details to capture hierarchical regional patterns, resulting in significantly improved similarity search performance.

For the trajectory classification task, BLUE’s improvement is less pronounced compared to other tasks, as classification is relatively easier and primarily depends on overall trajectory semantics (e.g., OD and travel patterns), which are less reliant on fine-grained details. Additionally, the Chengdu dataset features binary classification, resulting in consistently high performance across all methods, as reflected by the strong baseline accuracy.

\subsection{Micro Experiments}\label{exp:micro}
\noindent
\textbf{Ablation Study (RQ2).} To validate BLUE's designs, we experiment with 7 of its variants:
\begin{itemize}[leftmargin=*]
    \item \textbf{w/o P@2} removes the precision@2, the final trajectory representation is the [CLS] embedding of precision@3.
    \item \textbf{w/o P@3} removes the precision@3. The patch trajectory of precision@2 comes from precision@5. 
    \item \textbf{w/o P@5} removes the precision@5. We only remove the Transformer of precision@5 to keep the input as GPS trajectories. 
    \item \textbf{w/o Patch} removes patching module. We only keep the Transformer of level-1 and spatial-temporal reconstruction task.
    \item \textbf{w/ Min} replaces attention-pooling with min-pooling.
    \item \textbf{w/ Max} replaces attention-pooling with max-pooling.
    \item \textbf{w/ Mean} replaces attention-pooling with mean-pooling.
\end{itemize}

Table~\ref{tab:ablation} presents the results of the ablation study. For clarity, we report a single accuracy measure for each task. The results indicate that all designs contribute effectively to improving model performance. Overall, the impact on MSTS is most significant, as it directly relies on the quality of the pre-trained model. In contrast, tasks such as TTE and TC have less performance degradation due to fine-tuning. Specifically, the low-level information, i.e., P@5, has the most substantial impact on task performance, as it captures the essential spatial-temporal correlations within trajectories through the Transformer. Higher levels, on the other hand, have a relatively smaller effect on overall performance. This is because higher levels are primarily dependent on the lower-level module, and the trajectory’s semantic information is heavily compressed at these levels. Removing the patch module has little impact on TTE performance because GPS trajectories still retain the sampling rate information. However, it significantly degrades similarity performance, as GPS trajectories have low travel semantic density and are inherently irregular. In addition, different pooling strategies also influence the model performance. Min and Max pooling methods focus on extracting the most dominant features of the trajectory’s spatial-temporal information. Mean pooling retains more of the trajectory’s spatial-temporal integrity but can lead to feature smoothing, which may result in a loss of spatial-temporal details.

\begin{table}[!t]
   \LARGE
   \centering
   \caption{Accuracy for cross-dataset transfer, where the model is trained on one dataset and evaluated on another. For each method, the first row shows the accuracy on the target dataset, and the second row shows the relative change w.r.t. the source dataset. \textcolor{darkred}{Red} indicates an improvement over the no-transfer setting, while \textcolor{blue}{blue} denotes a performance drop. \colorbox{lightgray}{\textcolor{lightgray}{wo}} highlights the best relative change in transferability.
   }
   \resizebox{\linewidth}{!}{
   \begin{tabular}{l|ccc|ccc}\toprule
      & \multicolumn{3}{c}{Chengdu$\rightarrow$Porto}
      & \multicolumn{3}{c}{Porto$\rightarrow$Chengdu} 
      \\\cmidrule(lr){2-4}\cmidrule(lr){5-7}  
        & MAE$\downarrow$  & MR$\downarrow$ & Micro-F1$\uparrow$
        & MAE$\downarrow$  & MR$\downarrow$ & F1$\uparrow$ 
   \\
        & (TTE)  & (MSTS) & (TC)  
        & (TTE)  & (MSTS) & (TC)    \\
        \midrule
        \multirow{2}{*}{TrajCL}
        & 70.288 & 179.949 & 0.815 
	& 86.997 & 201.164 & 0.854    \\
        & \colorbox{lightgray}{\textcolor{blue}{8.58\%}} & \textcolor{blue}{4979\%} & \textcolor{blue}{0.97\%}    
 	& \textcolor{blue}{9.59\%} & \textcolor{blue}{3100\%} & \textcolor{blue}{2.06\%}  
        \\
        \midrule
        \multirow{2}{*}{START}
        & 95.234  & 262.764 & 0.793  
	& 94.583  & 302.430 & 0.866    \\
        & \textcolor{blue}{14.25\%} & \textcolor{blue}{8862\%} & \textcolor{blue}{3.17\%}  
 	& \textcolor{blue}{10.31\%} & \textcolor{blue}{8674\%} & \textcolor{blue}{2.04\%} 
        \\
        \midrule
        \multirow{2}{*}{MMTEC}
        & 74.213  & 102.433 & 0.808  
	& 82.533  & 113.324 & 0.838    \\
        & \textcolor{blue}{18.41\%} & \textcolor{blue}{4963\%} & \textcolor{blue}{4.83}\% 
 	& \textcolor{blue}{17.93\%} & \textcolor{blue}{4610\%} & \textcolor{blue}{1.30\%}
        \\
        \midrule
	\multirow{2}{*}{JGRM}
        & 70.493 & 91.453 & 0.802  
	& 86.212 & 120.065 & 0.824  \\
        & \textcolor{blue}{22.01\%} & \textcolor{blue}{4892\%} & \textcolor{blue}{1.35\%}
 	& \textcolor{blue}{29.73\%} & \textcolor{blue}{5885\%} & \textcolor{blue}{5.07\%}
        \\
        \midrule
	\multirow{2}{*}{\textbf{BLUE}} 
        & \textbf{7.556} & \textbf{1.063} & \textbf{0.825} 
	& \textbf{30.262} & \textbf{10.503} & \textbf{0.902}  \\
        & \textcolor{blue}{11.84\%} & \colorbox{lightgray}{\textcolor{darkred}{6.43\%}}  & \colorbox{lightgray}{\textcolor{blue}{0.96\%}} 
 	& \colorbox{lightgray}{\textcolor{blue}{3.28\%}} &\colorbox{lightgray}{\textcolor{blue}{629\%}} & \colorbox{lightgray}{\textcolor{blue}{0.22\%}} 
        \\
	\bottomrule
   \end{tabular}
   }
   \label{tab:transfer}
\end{table}

\stitle{Transferability Study (RQ3).} We conducted experiments to evaluate the transferability of BLUE compared to SOTA methods.  
The results, presented in Table~\ref{tab:transfer}, reveal significant performance degradation in all baseline models. This is primarily because road-based and grid-based methods rely on re-initializing grid IDs and road IDs when the road network scale or grid scale differs across cities, leading to reduced performance. Notably, for MSTS tasks that do not require fine-tuning, this re-initialization introduces a pronounced performance penalty. For BLUE, there is minimal performance loss across all tasks. This is because BLUE relies solely on unified normalized GPS and time information as input, without any external dependencies, granting the model strong versatility. Notably, BLUE can transfer effectively to tasks like MSTS that do not require fine-tuning. However, when transferring from Porto to Chengdu, there is a noticeable loss in MR due to Porto’s smaller city scale compared to Chengdu, resulting in data distributions that fail to cover the full range of Chengdu’s trajectories. Conversely, the transfer performance from Chengdu to Porto in MSTS surpasses the results without transfer, suggesting that models trained on larger cities may yield superior results when transferred to smaller cities.

To further explore the transferability of our model, we introduce a new dataset (Roma\footnote{https://ieee-dataport.org/open-access/crawdad-romataxi} with 90,172 trajectories) to test BLUE. Since Roma lacks classification labels, we evaluate the TTE and MSTS tasks. The dataset is split into train/val/test as [0.8, 0.1, 0.1], with 1,000 samples for the query/label set and the other 50,000 samples for the database set on the MSTS task. The results of Table~\ref{tab:transfer_roma} show that BLUE outperforms two SOTA baselines in both non-transfer and transfer settings. We also can observe that the accuracy of model transfer (from other cities to Roma) can be higher than the model trained on Roma. This is because the Chengdu and Porto datasets are much larger than the Roma dataset and thus allow for pre-training higher-quality models.

\begin{table}[!t]
\centering
\caption{Performance comparison of transferability on the Roma dataset with two SOTA methods.}
\label{tab:transfer_roma}
\resizebox{\linewidth}{!}{
\begin{tabular}{l l c c}
\hline
Method & Transfer Setting & MAE$\downarrow$(TTE) & MR$\downarrow$(MSTS) \\
\hline
TrajCL & No transfer         & 129.33 & 8.18 \\
       & Chengdu$\rightarrow$Roma        & 158.37 & 240.26 \\
       & Porto$\rightarrow$Roma          & 160.11 & 399.85 \\
\hline
JGRM   & No transfer         & 85.00  & 4.94 \\
       & Chengdu$\rightarrow$Roma        & 111.09 & 201.74 \\
       & Porto$\rightarrow$Roma          & 104.06 & 253.21 \\
\hline
\textbf{BLUE} & No transfer     & \textbf{57.22}  & \textbf{3.14} \\
           & Chengdu$\rightarrow$Roma    & \textcolor{darkred}{\textbf{52.58}}  & \textcolor{darkred}{\textbf{1.11}} \\
           & Porto$\rightarrow$Roma      & \textcolor{darkred}{\textbf{56.39}}  & \textcolor{darkred}{\textbf{1.98}} \\
\hline
\end{tabular}
}
\end{table}

\begin{figure}[!t]
    \centering
    \includegraphics[width=\linewidth]{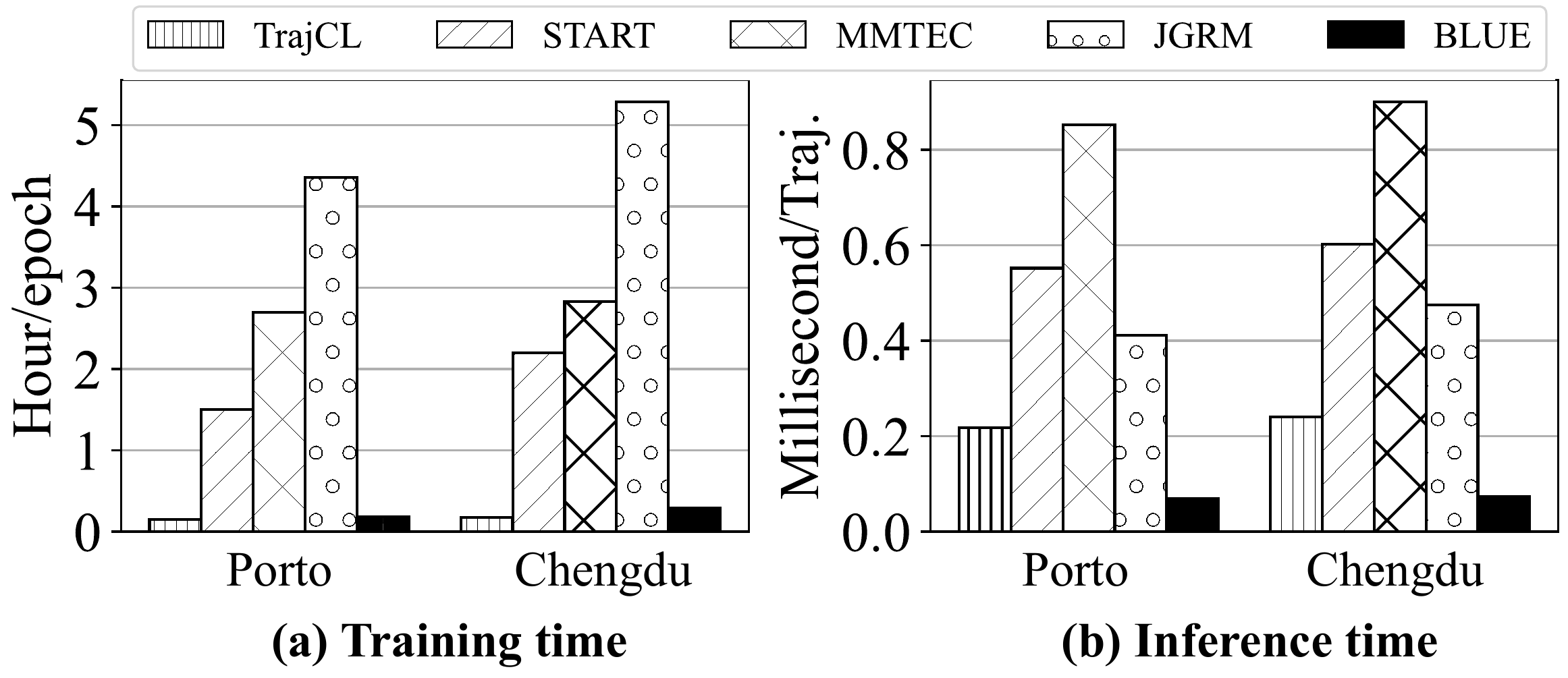}
    \caption{Training time (in hours) for one epoch and model inference time (in ms) of one trajectory on batch size 256.}
    \label{fig:efficiency}
\end{figure}

\stitle{Efficiency Study (RQ4).} We evaluate the efficiency of our model using three metrics: model size, training time, and inference time. The results, presented in Table~\ref{tab:model_size} and Figure~\ref{fig:efficiency}, demonstrate the high efficiency of BLUE. Since our model has no external dependencies, its size remains unaffected by the scale of the city. In contrast, the parameters of road-based and grid-based models increase substantially as city size grows. Despite incorporating 20 attention layers, our model achieves fast training times and efficient inference, largely due to its pyramid structure, which drastically reduces sequence lengths at higher levels. Notably, inference is exceptionally fast as it requires only the encoder. Additionally, the MSE loss in our model is more computationally efficient than the MLM loss with Softmax operations employed by START and JGRM. While TrajCL benefits from its lightweight, i.e., two-layer networks. MMTEC requires longer training times due to its reliance on NeuralCDE and its computationally complex network structure.

\stitle{Hyper-parameter Study (RQ5).} We also explore how the hyperparameters, including embedding dimensions and network layers, affect the performance of BLUE. Due to the page limit, we report the results in Appendix~\ref{sec:app_hyper}.

\begin{table}[!t]
\LARGE
    \centering
    \caption{The number of model parameters (M: million).}
    \label{tab:model_size}
    \resizebox{0.85\linewidth}{!}{
        \begin{tabular}{l|cccc|c}
        \toprule
        & TrajCL  & START & MMTEC & JGRM  & \textbf{BLUE}    \\
        \midrule
        Porto & 8.18M  & 8.10M & 1.65M & 10.13M   & \multirow{2}{*}{\textbf{3.51M}} \\
        \cmidrule{1-5}
        Chengdu & 28.16M  & 15.54M & 5.36M & 28.73M   & \\
        \bottomrule
        \end{tabular}
    }
\end{table}  

\section{Conclusion}
We propose BLUE, a self-supervised method for trajectory representation learning through varying GPS decimal precisions. BLUE is an encoder-decoder model with a pyramid structure. The encoder compresses GPS trajectories into patch trajectories from low-level to high-level, while the decoder restores GPS trajectories from high-level to low-level. Based on the pyramid structure, BLUE can learn hierarchical semantics of patch trajectories with multiple levels. Experimental results demonstrate that BLUE consistently outperforms existing TRL methods in both effectiveness and efficiency.

\section{Acknowledgement}
This paper was supported by the National Key R\&D Program of China 2024YFE0111800, and NSFC U22B2037, U21B2046, and 62032001.

\newpage
\bibliographystyle{ACM-Reference-Format}
\bibliography{ref}

\appendix
\newpage

\begin{figure}[!t]
    \centering
    \includegraphics[width=\linewidth]{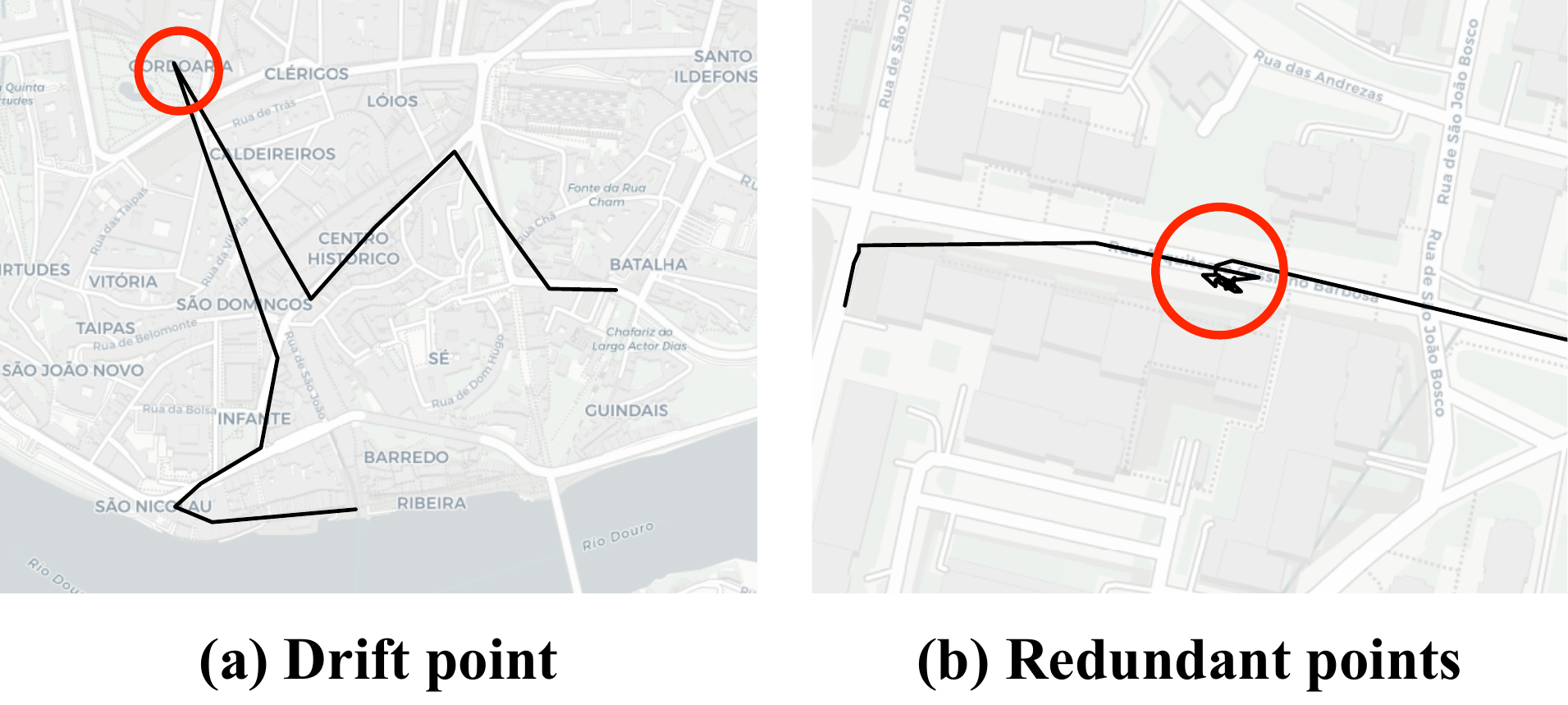}
    \caption{The illustration of drift and redundant 
 points.}
    \label{fig:preprocess}
\end{figure}

\section{Appendix}\label{sec:appendix}
\subsection{Algorithm} \label{sec:appendix_alg}
We provide an open-source pytorch-based implementation of blurred encoding. Specifically, this implementation avoids the for-loop operation in the patch encoder caused by the sequence that records all actual patch lengths of trajectories in a batch.

\begin{table}[tbp]
	\centering
	\caption{Average length of different types of trajectories.}
	\label{tab:length}
	\resizebox{0.9\linewidth}{!}{
		\begin{tabular}{l|c|c}\toprule
			\textbf{Trajectory Category} & Porto & Chengdu \\
			\midrule
                \textbf{Road traj. (\#Road segments)} & 27  & 28 \\
                \textbf{Grid traj. (\#Grid cells)} & 45    & 32 \\
                \textbf{Level-1 traj. (\#GPS points)} & 39 & 47 \\
                \textbf{Level-2 traj. (\#Patches)} & 27   & 31 \\
                \textbf{Level-3 traj. (\#Patches)} & 6     & 6 \\
			\bottomrule
		\end{tabular}
	}
\end{table}

\subsection{Complexity Analysis}\label{sec:complexity}
The time complexity of BLUE is mainly derived from the attention mechanism of the Transformer, which is given by $\mathcal{O}(n^2)$, where $n$ is the sequence length.
Suppose the patch trajectory lengths at level-1, level-2, and level-3 are $n_1$, $n_2$, and $n_3$, respectively, with corresponding Transformer layers $L_1$, $L_2$, and $L_3$, where $n_3 \ll n_2 \ll n_1$. The training time complexity of BLUE, including the patch encoder $\mathcal{O}(L_1 \cdot n_1^2 + L_2 \cdot n_2^2 + L_3 \cdot n_3^2)$ and the patch decoder $\mathcal{O}(L_1 \cdot n_1^2 + L_2 \cdot n_2^2 + L_3 \cdot n_3^2 + n_1\cdot n_2 + n_2 \cdot n_3 + n_2^2 + n_3^2)$, where $n_1\cdot n_2$ and $n_2 \cdot n_3$ are cross-attention, $n_2^2$ and $n_3^2$ are self-attention. The inference time complexity of BLUE, only including the patch encoder, is $\mathcal{O}(L_1 \cdot n_1^2 + L_2 \cdot n_2^2 + L_3 \cdot n_3^2)$. Due to the reduction in trajectory length, the actual operation becomes highly efficient, i.e., approximates to $\mathcal{O}(L_1 \cdot n_1^2)$. Table~\ref{tab:length} in Appendix presents the average sequence lengths of different types of trajectories. The average sequence length of the patch trajectories decreases significantly, with the average length at level-3 being a single digit.

\subsection{Data Preprocessing} \label{sec:appendix_preprocessing}
Like previous studies~\cite{start,jgrm}, we preprocess datasets as follows: 1) we remove trajectories less than 1 kilometer in length. 2) For the Chengdu dataset, we sample a sub-dataset. We also remove patch trajectories at precision@2 whose sequence lengths are less than 2 to ensure that patch trajectories at precision@2 are valid sequences. Moreover, considering the noise of the GPS trajectory, we also do the following processing: 1) Delete drift points. Drift points, as shown in Figure~\ref{fig:preprocess}(a), cause significant interference with the travel semantics, and also make the wrong road trajectories and grid trajectories for baselines. We remove the drift points directly based on the abnormal driving speed\footnote{https://github.com/ni1o1/transbigdata}. 2) Redundant points, as illustrated in Figure~\ref{fig:preprocess}(b), refer to dozens or even hundreds of GPS points clustered within a very small area, resulting in significant information redundancy and unnecessarily extending the GPS trajectory sequence. This redundancy hinders effective model learning. Notably, both grid cells and road segments provide efficient solutions to this issue. Grid cells aggregate these clustered points into a grid cell, while road segments align them to a road segment, thereby reducing redundancy. To address this issue in GPS trajectories, we take a radius of 50 meters of GPS points as a small region, and if there are more than 10 points in the region, we only keep the first point and the last point\footnote{https://github.com/movingpandas/movingpandas} to reduce redundancy. The timestamps of these two points can retain information such as congested segments.

\subsection{Baselines} \label{sec:appendix_baselines}
\noindent
\textbf{GPS-based Methods:}
\begin{itemize}[leftmargin=*]
    \item \textbf{Traj2vec}~\cite{traj2vec}\footnote{https://github.com/yaodi833/trajectory2vec}: An RNN-based seq2seq model converts GPS trajectory into GPS feature sequence to learn trajectory representations for the trajectory clustering task.
\end{itemize}

\noindent
\textbf{Grid-based Methods:}
\begin{itemize}[leftmargin=*]
    \item \textbf{TrajCL}~\cite{trajcl}\footnote{https://github.com/changyanchuan/TrajCL}: The state-of-the-art grid-based method proposes a set of trajectory augmentations on grid trajectory in free space and dual-feature self-attention to learn grid trajectory representations using contrastive learning with the Transformer. We set grid size as 100m$\times$100m.
\end{itemize}

\noindent
\textbf{Road-based Methods:}
\begin{itemize}[leftmargin=*]
    \item \textbf{Trembr}~\cite{trembr}: An RNN-based seq2seq model uses spatial-temporal road trajectory sequences to learn trajectory representation.
    \item \textbf{PIM}~\cite{pim}\footnote{https://github.com/Sean-Bin-Yang/Path-InfoMax}: It first uses node2vec~\cite{node2vec} to learn road embeddings on the road network and then proposes mutual information maximization to learn trajectory representations.
    \item \textbf{JCLRNT}~\cite{JCLRNT}\footnote{https://github.com/mzy94/JCLRNT}: It uses GAT and Transformer to learn road representations and trajectory representations. Three different contrastive losses are designed to optimize the model.
    \item \textbf{START}~\cite{start}\footnote{https://github.com/aptx1231/START}: This method uses BERT~\cite{bert} as the trajectory encoder that integrates travel semantics and temporal regularities with contrastive learning and two self-supervised tasks.
\end{itemize}

\noindent
\textbf{Multi-modal Methods:}
\begin{itemize}[leftmargin=*]
    \item \textbf{MMTEC}~\cite{mmtec}\footnote{https://github.com/Logan-Lin/MMTEC}: The multi-modal method learns trajectory representations using an attention-based discrete encoder and a NeuralCDE continuous encoder that extract travel behavior and continuous spatial-temporal correlations of trajectories.
    \item \textbf{JGRM}~\cite{jgrm}\footnote{https://github.com/mamazi0131/JGRM}: This multi-modal method learns trajectory representations both in the view of free-space and the view of the road network space jointly with three different loss functions, i.e., contrastive loss, MLM loss, and alignment loss.
\end{itemize}

\subsection{Downstream Tasks Settings}\label{sec:appendix_downstream settings}
\noindent
\textbf{Travel Time Estimation.} Travel time estimation is a fine-tuning task that aims to predict the travel time of a trajectory. This task focuses on temporal information and sampling rate information.
We add a two-layer MLP after the patch encoder to predict travel time on \textit{seconds} and use mean square error as a loss function. Since inputs of the pre-training model have full-time information, we only retain the start time and ignore other times to avoid information leakage in fine-tuning~\cite{start}. We set the learning rate to 1e-4 and other hyperparameters to be consistent with pre-training.

\stitle{Trajectory Classification.} Trajectory classification is also a fine-tuning task, which assigns a trajectory to a category. This task needs the pattern of movement. We classify trajectories according to the travel mode in Porto, which is a multi-classification. We determine if there are passengers in the taxi of Chengdu, which is a binary classification. We add a two-layer MLP after the patch encoder to predict the category label and use cross-entropy as a loss function. We set the learning rate to 1e-4 and other hyperparameters to be consistent with pre-training.

% \balance

\begin{figure}[!t]
    \centering
    \includegraphics[width=\linewidth]{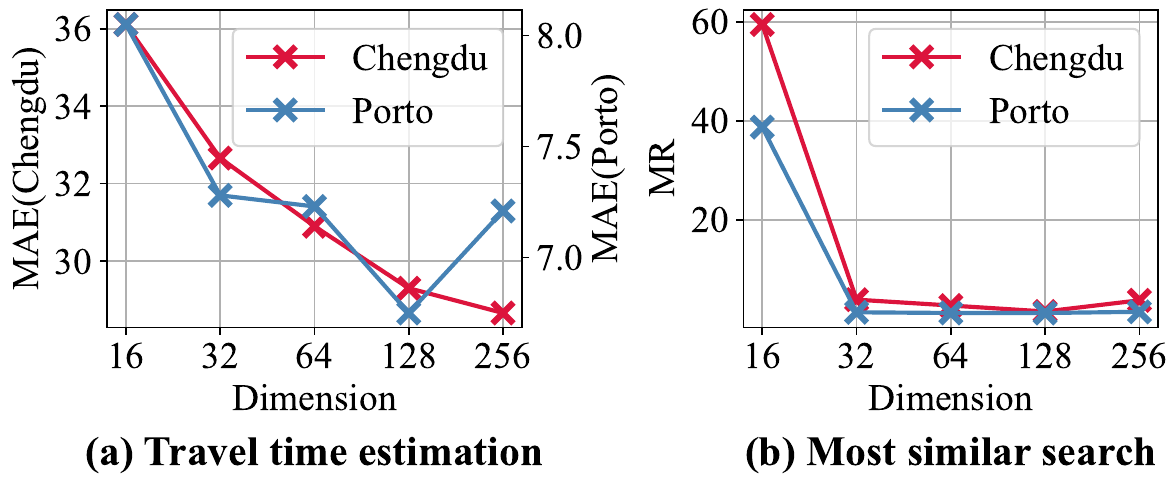}
    \caption{Effect of embedding dimension.}
    \label{fig_exp:dimension}
\end{figure}

\begin{figure}[!t]
    \centering
    \includegraphics[width=\linewidth]{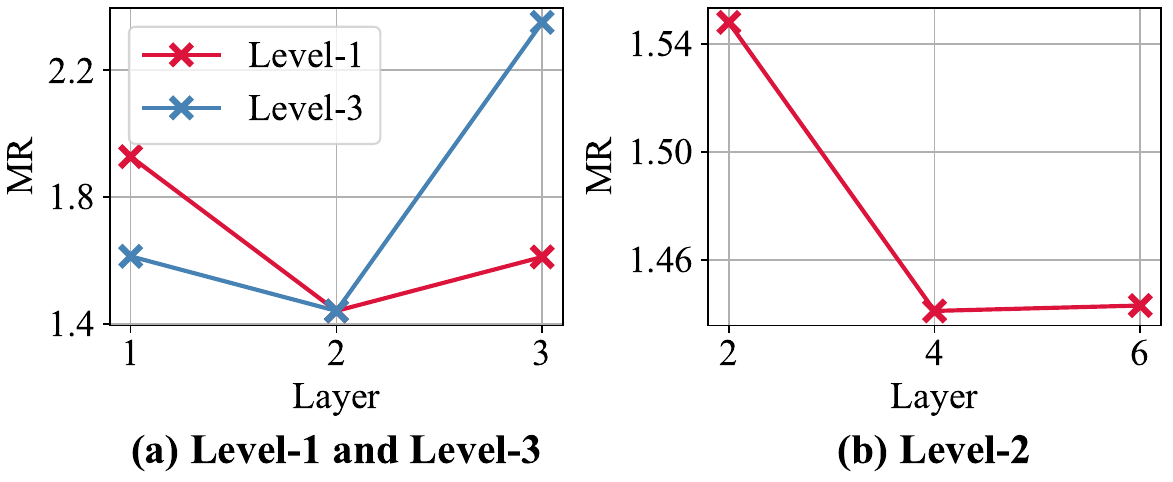}
    \caption{Effect of layers in Chengdu on similarity.}
    \label{fig_exp:layer}
\end{figure}

\stitle{Most Similar Trajectory Search.} Given a query trajectory $T$ of the query dataset $\mathcal{D}_q$, the most similar trajectory search is to find the most similar trajectory $T^\prime$ from a large trajectory database $\mathcal{D}_d$. This task relies on spatial information (e.g., trajectory shape and origin-destination). We directly use pre-trained trajectory representations of the patch encoder with dot product of vectors to evaluate the performance (No fine-tuning). However, the lack of ground truth makes it difficult to evaluate the accuracy. Followed by TrajCL~\cite{trajcl}, we randomly select 1,000 trajectories as the query set $\mathcal{D}_q$ and another 100,000 trajectories as the database $\mathcal{D}_d$. Then, we keep the start and end GPS points unchanged to keep their overall origin-destination (OD) similar, and downsample (ratio 0.3) other GPS points in each trajectory of $\mathcal{D}_q$ to obtain sub-trajectories as the most similar trajectories set $\mathcal{D}_q^\prime$. We add $\mathcal{D}_q^\prime$ to $\mathcal{D}_d$ to obtain the final database set $\mathcal{D}_d^\prime$. For baselines, we directly transform sub-trajectories to sub-grid trajectories for grid-based methods. For road-based methods, the influence of downsampling can be reduced by map matching on road trajectories~\cite{start}. Therefore, to keep the continuity of road trajectories, we adopt the detour method~\cite{JCLRNT}, which involves randomly selecting a middle continuous road segment constituting 30\% of the length of the road trajectory and replacing it with an alternative detour.

\subsection{The Effect of Hyper-parameters}\label{sec:app_hyper}
\stitle{Dimension.} We experiment with different hidden embedding dimensions for the final trajectory representation, ranging from $[16, 32, 64, 128, 256]$. Figure~\ref{fig_exp:dimension} shows the results for two tasks: travel time estimation (fine-tuning) and most similar trajectory search (no fine-tuning) on two datasets. In general, both tasks improve with a larger embedding dimension, as it allows for more comprehensive encoding of travel semantics in the hidden embedding and the final trajectory representation. Specifically, when the Chengdu dataset increases from 128 to 256 dimensions, travel time estimation accuracy improves, while for the Porto dataset, accuracy decreases. This suggests that dimensional trends vary on different datasets, but overall, the 128-dimensional embedding achieves the best balance between performance and efficiency.

\stitle{Layer.} We vary the number of layers at each level: for level-1 and level-3, the number of layers changes between [1, 2, 3], and for level-2, it changes between [2, 4, 6]. We conducted experiments on the Chengdu dataset, and the results are shown in Figure~\ref{fig_exp:layer}. The overall trends for level-1 and level-3 are similar. When the number of layers is set to 1 or 3, performance decreases. This happens because with too few layers, the trajectory information cannot be effectively learned. For level-1, due to the limited amount of GPS trajectory information, having too many layers can lead to overfitting. For level-3, it also does not need too many layers, because the length of the patch sequence is greatly reduced. However, level-2 requires more layers since it serves as the middle layer, built from level-1 and supporting level-3. Therefore, it needs deeper modules for effective representation. Notably, the performance of level-3 stabilizes when the number of layers is set to 4.

% \end{sloppypar}
\end{document}